# Augmentation of the Reconstruction Performance of Fuzzy C-Means with an Optimized Fuzzification Factor Vector

Kaijie Xu, Witold Pedrycz, *Fellow, IEEE*, and Zhiwu Li, *Fellow, IEEE*

*Abstract*—Information granules have been considered to be the fundamental constructs of Granular Computing (GrC). As a useful unsupervised learning technique, Fuzzy C-Means (FCM) is one of the most frequently used methods to construct information granules. The FCM-based granulation-degranulation mechanism plays a pivotal role in GrC. In this paper, to enhance the quality of the degranulation (reconstruction) process, we augment the FCM-based degranulation mechanism by introducing a vector of fuzzification factors (fuzzification factor vector) and setting up an adjustment mechanism to modify the prototypes and the partition matrix. The design is regarded as an optimization problem, which is guided by a reconstruction criterion. In the proposed scheme, the initial partition matrix and prototypes are generated by the FCM. Then a fuzzification factor vector is introduced to form an appropriate fuzzification factor for each cluster to build up an adjustment scheme of modifying the prototypes and the partition matrix. With the supervised learning mode of the granulation-degranulation process, we construct a composite objective function of the fuzzification factor vector, the prototypes and the partition matrix. Subsequently, the particle swarm optimization (PSO) is employed to optimize the fuzzification factor vector to refine the prototypes and develop the optimal partition matrix. Finally, the reconstruction performance of the FCM algorithm is enhanced. We offer a thorough analysis of the developed scheme. In particular, we show that the classical FCM algorithm forms a special case of the proposed scheme. Experiments completed for both synthetic and publicly available datasets show that the proposed approach outperforms the generic data reconstruction approach.

## I. INTRODUCTION

Granular Computing (GrC) is a computation paradigm and emerging conceptual framework of information processing, which plays an important role in many areas [1]. Information granules are considered to be the fundamental building blocks of GrC. As a useful unsupervised learning technique, fuzzy clustering has become a powerful approach to information granulation [2]. It offers a suite of algorithms aiming at the discovery of a structure in the given dataset [3]. This type of approach partitions a given input space into several regions, depending upon some preselected similarity measures. One of the most widely used and effective fuzzy clustering approaches is Fuzzy C-Means (FCM) [4]. It has been experimentally demonstrated that quite commonly this algorithm improves the performance of classification compared with the traditional clustering algorithms.

From the general viewpoint, the FCM is usually regarded as a granular information technique, where information granule is represented by its prototype (center of the cluster) and a partition matrix. Both descriptors are numeric. With the aid of constructed prototypes and partitions [5], data are encoded into information granules. In other words, numeric data are described as prototypes and partition matrices, which is the so-called granulation mechanism. Clustering granular instead of numeric data provides a novel and interesting avenue of investigation [6]. Cluster prototypes and partition matrices are obtained by optimizing the fuzzy set-based clustering model [7]. In the FCM-based granulation progress, fuzzy clustering approaches are used to cluster the numerical data into fuzzy information granules [8-15]. Degranulation, as an inverse problem of granulation that involves the reconstruction of numeric results on the basis of already constructed information granules, is also task worth studying. The reconstruction is usually referred to as a degranulation or decoding process. To some extent degranulation can also reflect the performance of granulation mechanism (classification performance of the fuzzy clustering) [2, 5].

The mechanism of granulation-degranulation [16] involves a series of processes of dealing with fuzzy information granules. It plays an important role in GrC, just as analog-to-digital (A/D) conversion as well as digital-to-analog (D/A) conversion in the field of signal processing [17], and fuzzification-defuzzification in the field of fuzzy control systems. The classification rate and reconstruction (degranulation) error are often used as the performance evaluation indexes of the FCM-based granulation-degranulation mechanism. Previous studies indicate that the reconstruction (degranulation) and the classification (granulation) are related to each other [4]. In most cases, the smaller the degranulation error is, the better the performance of granulation becomes [4].

In [18], the reconstructed data supervised by the original data

This work was supported in part by the National Natural Science Foundation of China under Grant Nos. 61672400 and 61971349. (*Corresponding author: Zhiwu Li*).

K. Xu is with the School of Electro-Mechanical Engineering, Xidian University, Xi'an 710071, China, Department of Electrical and Computer Engineering, University of Alberta, Edmonton, AB T6R 2V4, Canada and also with the Department of remote sensing science and technology, School of Electronic Engineering, Xidian University, Xi'an 710071, China (e-mail: kjxu@stu.xidian.edu.cn).

W. Pedrycz is with the Department of Electrical and Computer Engineering, University of Alberta, Edmonton, AB T6R 2V4, Canada, the School of Electro-Mechanical Engineering, Xidian University, Xi'an 710071, China, Systems Research Institute, Polish Academy of Sciences, Warsaw, Poland and also with the Faculty of Engineering, King Abdulaziz University, Jeddah 21589, Saudi Arabia (e-mail: wpedrycz@ualberta.ca).

Z. Li is with the School of Electro-Mechanical Engineering, Xidian University, Xi'an 710071, China, and also with the Institute of Systems Engineering, Macau University of Science and Technology, Macau 999078, China (e-mail: zhwli@xidian.edu.cn).

is introduced into the FCM clustering, which makes the performance of fuzzy clustering enhanced. To improve the quality of degranulation, in [5] Hu *et al.* use a linear transformation of the membership matrix to refine the prototypes, making the granulation-degranulation mechanism optimized. In [19], Rubio *et al.* combine the GrC and granulation-degranulation mechanism to design a granular-based FCM algorithm that makes the numerical data more reflective.

Hitherto, the mechanism of granulation-degranulation has not been widely and deeply touched upon. The lack of a well-established body of knowledge breaks up new opportunities and calls for more investigations in this area.

The main objective of the paper is to develop an enhanced scheme of data reconstruction to improve the performance of the degranulation mechanism. In the proposed scheme, the notion of fuzzification factor vector is introduced such that we can assign an appropriate fuzzification factor to each prototype. Then, an adjustment mechanism of the prototype and the partition matrix is established in the supervised learning mode of the granulation-degranulation mechanism. Subsequently, particle swarm optimization (PSO) [20] is used to determine an optimal fuzzification factor vector by minimizing the degranulation error. Thus, with the optimal fuzzification factor vector, a reasonable partition matrix and a collection of the optimized prototypes are obtained. Finally, the performance of degranulation mechanism (reconstruction) is enhanced. The augmented granulation-degranulation mechanism can obtain superior quality reconstructed data through the modified partition matrix and the refined prototypes. Both the theoretical investigations and experimental results demonstrate that the proposed scheme outperforms the generic way of data degranulation. To the best of our knowledge, the idea of the proposed approach has not been exposed in previous studies.

This paper is organized as follows. A generic way of data granulation and degranulation is briefly reviewed in Section II. The theory of the proposed scheme is introduced in Section III. Section IV includes experimental setup and the analysis of experiments. Section V concludes this paper.

## II. Granulation-Degranulation Mechanism

In this part, we briefly review the FCM-based granulation-degranulation process.

### A. Granulation

With an FCM algorithm, the structure in a dataset $X$ ($X \in R^n$) is expressed in terms of prototypes (clusters) and a partition matrix. Then, the dataset is encoded into fuzzy information granules with the aid of the constructed prototypes and the partition matrix, which is the called granulation. The FCM algorithm minimizes the distance-based cost function [21-22]:

$$J = \sum_{i=1}^{N}\sum_{j=1}^{C}\mu_{ij}^m d_{ij}^2 = \sum_{i=1}^{N}\sum_{j=1}^{C}\mu_{ij}^m \|x_i - v_j\|^2$$
$$x_i = [x_1, x_2\cdots, x_k, \cdots] \in R^{1\times n}, \ X = [x_1; x_2; \cdots; x_i; \cdots] \in R^{N\times n}$$
$$i=1,2,\cdots,N, \ k=1,2,\cdots,n, \ j=1,2,\cdots,C \quad (1)$$
s.t.
$$\sum_{j=1}^{C}\mu_{ij} = 1, \ 0 < \sum_{i=1}^{N}\mu_{ij} < N$$

where $x_i$ ($i=1,2,\cdots,N$) is the $i$th data of $X$, $v_j$ ($j=1,2,\cdots,C$) is the $j$th prototype (center) of the cluster, $\mu_{ij}$ is the degree of membership of the $i$th data belonging to the $j$th prototype, and $m(m>1)$ is a scalar representing the fuzzification factor (coefficient) that exhibits a significant impact on the form of the developed clusters [17]. $\|\cdot\|$ stands for a certain distance function (in this paper, the weighted Euclidean distance is used) [17]. The above cost function is minimized by iteratively updating the partition matrix $U$ and the prototype matrix $V$ [23], i.e.,

$$U = [U_1, U_2, \cdots, U_i, \cdots, U_N] \in R^{C\times N}$$
$$U_i = [\mu_{i1}, \mu_{i2}, \cdots, \mu_{ij}, \cdots, \mu_{iC}]^T \quad (2)$$
$$\mu_{ij} = \left[\sum_{k=1}^{C}(\frac{\|x_i - v_j\|}{\|x_i - v_k\|})^{\frac{2}{m-1}}\right]^{-1}$$

$$V = \begin{bmatrix} v_1 \\ v_2 \\ \vdots \\ v_j \\ \vdots \\ v_C \end{bmatrix} \quad v_j = \frac{\sum_{i=1}^{N}(x_i \mu_{ij}^m)}{\sum_{i=1}^{N}\mu_{ij}^m} \quad (3)$$

where $T$ stands for the transpose operation [24]. Thus, with the FCM, the dataset $X$ is expressed as the prototype matrix $V$ and the partition matrix $U$.

### B. Degranulation

As an inverse problem of granulation, the degranulation involves the reconstruction of numeric results on the basis of already constructed information granules, more specifically the prototypes of the clusters. The form of the degranulation formula results from the minimization of the objective function:

$$J(x) = \sum_{i=1}^{N}\sum_{j=1}^{C}\mu_{ij}^m d_{ij}^2 = \sum_{i=1}^{N}\sum_{j=1}^{C}\mu_{ij}^m \|x_i - v_j\|^2 \quad (4)$$

Using the method of Lagrange multipliers, we find a minimum of the objective function, the reconstructed data can be solved as follows:

$$J(x,\lambda) = \sum_{i=1}^{N}\sum_{j=1}^{C}\mu_{ij}^m \|x_i - v_j\|^2 + \lambda \sum_{i=1}^{N}\left(\sum_{j=1}^{C}\mu_{ij} - 1\right) \quad (5)$$

where $\lambda$ is the Lagrange multiplier. We determine partial derivative of $x_i$ with respect to $J_\lambda$ and make it equal to zero. The solution to the reconstruction problem comes as follows:

$$\nabla J_v = 2\sum_{j=1}^{C}\mu_{ij}^m (x_i - v_j) = 0 \quad (6)$$

$$\hat{x}_i = \frac{\sum_{j=1}^{C}\mu_{ij}^m v_j}{\sum_{j=1}^{C}\mu_{ij}^m} \quad \hat{X} = [\hat{x}_1; \hat{x}_2; \cdots; \hat{x}_i; \cdots; \hat{x}_N]^T \quad (7)$$

It can be seen from (7) that each prototype is weighted by the corresponding coordinates of $\mu$, and the fuzzification factor is an integral part of this aggregation of the prototypes.

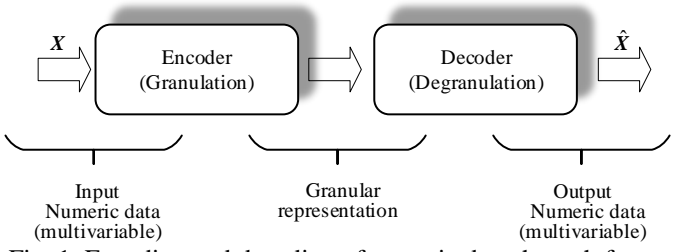

Fig. 1. Encoding and decoding of numeric data through fuzzy information granules.

The FCM-based granulation-degranulation mechanism can be organized in the two phases [17] as displayed in Fig. 1.

## III. An Enhanced Scheme of Degranulation Mechanism

As mentioned above, the fuzzification factor is an important parameter in FCM-based granulation-degranulation mechanism, which exhibits a significant impact on the form of the developed clusters. In the existing methods the fuzzification factor is usually set as a numerical value (a scalar) for a dataset to adjust (control) the resulting prototypes and indirectly affect the partition matrix. However, for most datasets the contribution of each datum to the prototypes is significantly different. In other words, the performance of the degranulation can be enhanced by setting an appropriate fuzzification factor for each prototype. In this section we build a supervised model of granulation-degranulation mechanism and optimize it to improve the performance of degranulation mechanism.

In this study, the degranulation error is quantified the following expression

$$R_e = \frac{1}{N} \|\hat{X} - X\|_2 \quad (8)$$

where $\|A\|_2$ is the 2-norm [25], expressed as

$$\|A\|_2 = \sqrt{\lambda_{\max}} \quad (9)$$

where $\lambda_{\max}$ is the largest eigenvalue of $A^H A$, and $H$ represents the conjugate transpose [26]. Obviously, the value of $R_e$ is affected by the number of clusters $C$ and the fuzzification factor $m$. Next, we concentrate on the detailed realization of the augmentation of degranulation mechanism.

### A. Construction of an Objective Function of a Fuzzification Factor Vector

In order to facilitate the analysis and ensuing design, two expressions of granulation-degranulation mechanism are established:

$$V = \Gamma X = \Phi U^m X \quad (10)$$

$$\hat{X} = \Omega V = \Theta [U^m]^T V \quad (11)$$

where $\Phi$ and $\Theta$ are two diagonal matrices coming in the following form:

$$\Phi = diag\left\{\frac{1}{\sum_{i=1}^{N} \mu_{i1}^{m_1}}, \cdots, \frac{1}{\sum_{i=1}^{N} \mu_{ij}^{m_j}}, \cdots, \frac{1}{\sum_{i=1}^{N} \mu_{iC}^{m_C}}\right\} \in R^{C \times C} \quad (12)$$

$$\Theta = diag\left\{\frac{1}{\sum_{j=1}^{C} \mu_{1j}^{m_j}}, \cdots, \frac{1}{\sum_{j=1}^{C} \mu_{ij}^{m_j}}, \cdots, \frac{1}{\sum_{j=1}^{C} \mu_{Nj}^{m_j}}\right\} \in R^{N \times N} \quad (13)$$

where $m = [m_1, \cdots, m_j, \cdots]$ is a fuzzification factor vector. The dimensionality of $U^m$ is $C \times N$ and its elements take the form

$$\mu_{ij}^{m_j} = \left[\sum_{k=1}^{C} \left(\frac{\|x_i - v_j\|}{\|x_i - v_k\|}\right)^{\frac{2}{m_j - 1}}\right]^{-m_j} \quad (14)$$

$$i = 1, 2, \cdots, N; \ j = 1, 2, \cdots, C$$

It is clear that in the generic way of data granulation and degranulation all the elements of the fuzzification factor vector are equal $(m_1 = m_2 = \cdots = m_j)$. To obtain a sound vector of fuzzification factors, we build and optimize a composite objective function in the form:

$$V(m) = diag\left\{\frac{1}{\sum_{i=1}^{N} \mu_{i1}^{m_1}}, \cdots, \frac{1}{\sum_{i=1}^{N} \mu_{ij}^{m_j}}, \cdots, \frac{1}{\sum_{i=1}^{N} \mu_{iC}^{m_C}}\right\} U^m X \quad (a)$$

$$f(V, U^m) = \|\Theta [U^m]^T V - X\|_2 \quad (b) \quad (15)$$

We have to determine the optimal fuzzification factor vector to refine the prototypes and optimize the partition matrix so as to minimize the objective function $f$.

### B. Optimization of the Composite Objective Function with Particle Swarm Optimization (PSO)

The model proposed in this study consists of two stages: an unsupervised clustering (granulation mechanism) stage, and a supervised refinement stage (supervised by the degranulation mechanism).

At the first stage, a dataset $X$ is granulated into the partition matrix $U$ and the prototypes $V$ with the FCM algorithm. To further improve the performance of degranulation (data reconstruction), at the second stage, a reconstruction criterion is introduced to supervise and refine the prototypes to determine a more reasonable partition matrix. Thus, the performance of degranulation is enhanced.

More specifically, we first set an initial fuzzification factor vector $m_0$ for the partition matrix to form the prototypes a disturbance according to (15-a). The modified prototypes are expressed as follows:

$$\hat{V} = V(m_0) \quad (16)$$

Then, with the new prototypes we modify the partition matrix in the following way

$$[\hat{U}^m]^T = diag\left\{\cdots, \left[\sum_{k=1}^{C}\left(\frac{1}{\|x_i - \hat{v}_k\|}\right)^{\frac{2}{m_j-1}}\right]^{-m_j}, \cdots\right\}\left[\cdots, \|x_i - \hat{v}_j\|^{2\frac{-m_j}{m_j-1}}, \cdots\right]$$

$$j = 1, 2, \cdots, C; \ i = 1, 2, \cdots, N \quad (17)$$

Subsequently, we calculate the reconstructed dataset (degranulation) $\hat{X}$ according to (11) with the new partition matrix $\hat{U}$ and the prototypes $\hat{V}$. In the process of granulation and degranulation, we use $R_e$ as a performance index to optimize the fuzzification factor vector to determine an acceptable fuzzification factor vector.

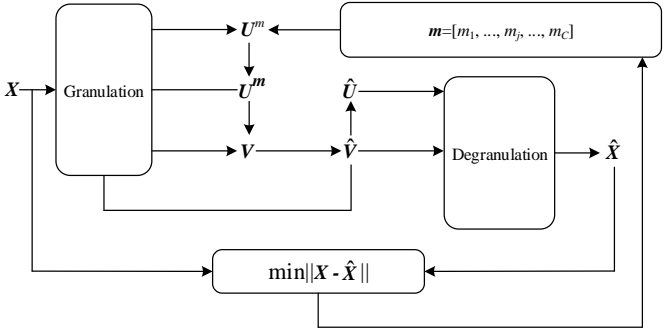

Fig. 2. An overall model: main processing phases.

Obviously, the composite objective function leads to a possible multimodality problem, which is not effective enough to be solved with the traditional optimization approaches [20]. However, population-based optimization algorithms are promising alternatives. In this paper, we use particle swarm optimization (PSO) [27, 28] as an optimization vehicle. Fig. 2 shows the flow of processing completed by the proposed scheme.

## IV. EXPERIMENTAL STUDIES

A series of experiments involving a synthetic dataset and several publicly available datasets [29, 30] coming from the machine learning repository are reported. The objective of the experiments is to compare the reconstruction performance of the proposed method with the FCM method. All data are normalized to have zero mean and unit standard deviation.

In the experiments, several different values of fuzzification coefficients and the number of clusters are considered. For each dataset the number of clusters is taken from 2 to 6, and the fuzzification coefficient $m$ is taken from 1.1 to 5.1, with a step size of 0.5. For the proposed method, the initial fuzzification factor vector is set as $[m, m, …, m]$. The algorithms are repeated 10 times and the means and standard deviations of the experimental results are presented. The algorithms terminate once the following termination condition has been met:

$$\max\left(\|U - U'\|\right) \leq 10^{-5} \quad (18)$$

where $U'$ is the partition matrix obtained at the previous iteration.

In the proposed method, the PSO algorithm [31, 32] is used to optimize the fuzzification factor vector in the $C$-dimensional space, where the number of the particle is $N$, $X_i = [X_{i1}, X_{i2} \cdots, X_{iC}]$ is the position vector of particle $i$, its velocity is $V_i = [V_{i1}, V_{i2} \cdots, V_{iC}]$, and its best position vector is $P_i = [P_{i1}, P_{i2} \cdots, P_{iC}]$. $P_g = [P_{g1}, P_{g2} \cdots, P_{gC}]$ is the best position vector of all particles. Each particle velocity is updated in the following way:

$$V_i^{n+1} = V_i^n + c_1 r_1 \left(P_i - X_i^n\right) + c_2 r_2 \left(P_g - X_i^n\right) \quad (19)$$

Then each particle position vector is updated by

$$X_i^{n+1} = X_i^n + V_i^n \quad (20)$$

In (19) and (20), $c_1$ and $c_1$ are cognitive weights, $r_1$ and $r_1$ are inertia weights drawn from [0, 1], and $n$ denotes evolutionary epochs [32].

The steps of the algorithm are outlined as follows:

1. Initialize the parameters of the algorithm, including the population size $N$, inertia weight, cognitive weight, social weight, $p_{best}$, $g_{best}$, etc [33].
2. Calculate the fitness of each particle according to (15-b).
3. Calculate $p_{best}$ for each particle and update $g_{best}$ for the swarm.
4: Update the velocity and the position of each particle according to (19) and (20);
5: Repeat the above process (Step 2 to Step 4) until the termination condition has been reached. If the preconditions are met, then stop iteration, output the optimal solution.

In the implementation of the PSO, the method is run for $t_{max} = 500$ iterations with 75 particles; however, we allow the algorithm to be terminated if no changes in $g_{best}$ [33] in $15\% t_{max}$ consecutive iterations. The other parameters of PSO are listed below: inertia weight=0.8, cognitive weight=1.49445 and social weight=1.49445 which are by far the most commonly used values [34]. To estimate the effectiveness of the proposed approach, we use a 5-fold cross validation [4, 35], which is commonly used to estimate (and validate) the performance of granulation-degranulation models.

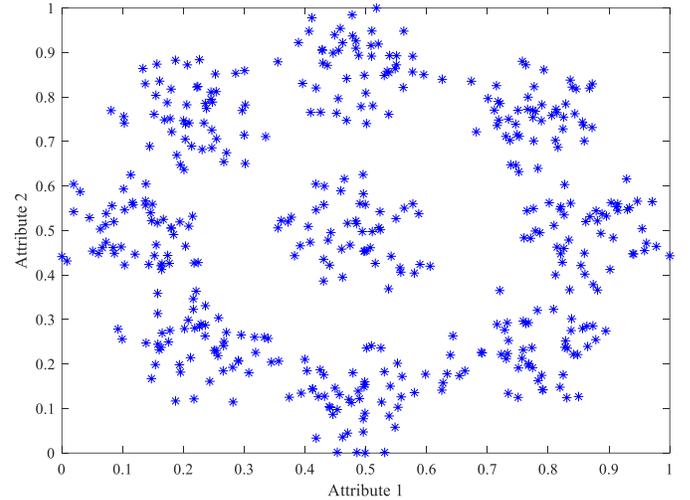

Fig. 3. Synthetic dataset.

*A. Synthetic data*

First, we report the results of reconstruction performance for an illustrative 2-D synthetic dataset, with a number of individuals 450 and nine categories in detail. The geometry of the dataset is visualized in Fig. 3. The optimization of the reconstruction error and the corresponding fuzzification factor vector with PSO are plotted in Figs. 4 and 5, respectively. Figs. 6 and 7 show the membership functions and their contour plots. It can be seen from Fig. 4 that the reconstruction error and the objective function decrease as the iterative process proceeds, which indicates that assigning an appropriate weight to each prototype can enhance the performance of the degranulation mechanism. In particular, the values of the starting position of two curves are obtained by using the FCM algorithm since we set the initial fuzzification factor vector as $[m, m, …, m]$ when using PSO. In addition, it also reflects that the objective function decreases with the reduction of reconstruction error.

In the iterative process, the elements of the fuzzification factor vector become scatter and finally reach a steady-state.

Ultimately, the membership functions are optimized (the curves become smooth) with the fuzzification factor vector, as shown in Figs. 6 and 7. It can be seen from fig. 7 that with the proposed method, the fuzzy membership values around the prototypes are enlarged. In other words, the contributions of these data to the prototypes are enhanced, which also indirectly makes the other ones reduced.

Then, in Figs. 8–10 we plot the results of the reconstruction errors (mean and standard deviation) of 5-fold cross validation with all the protocols [2] for the synthetic dataset and their corresponding fuzzification factors. Obviously, by assigning a reasonable fuzzification factor to each prototype, the reconstruction errors of the training set and the testing set are reduced.

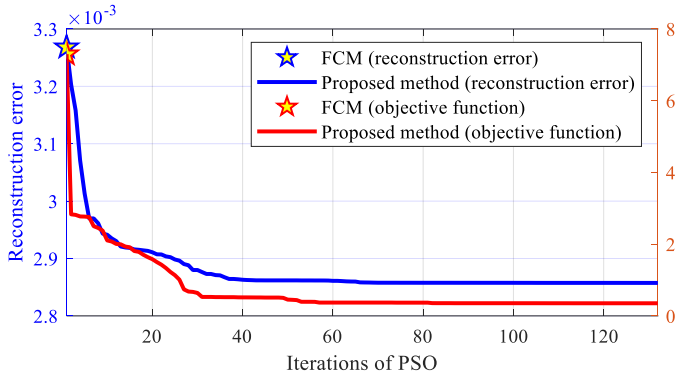

Fig. 4. Optimization of the reconstruction error with PSO.

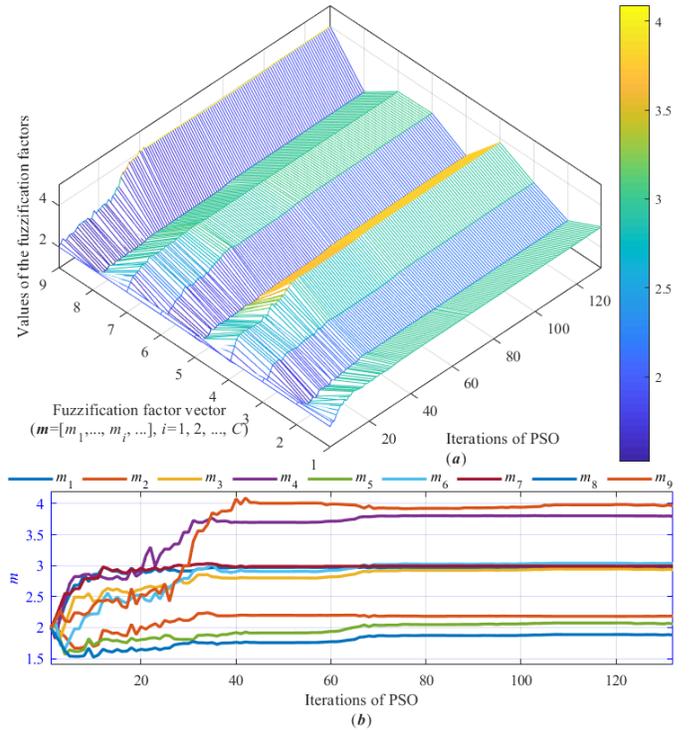

Fig. 5. The fuzzification factor vector with PSO.

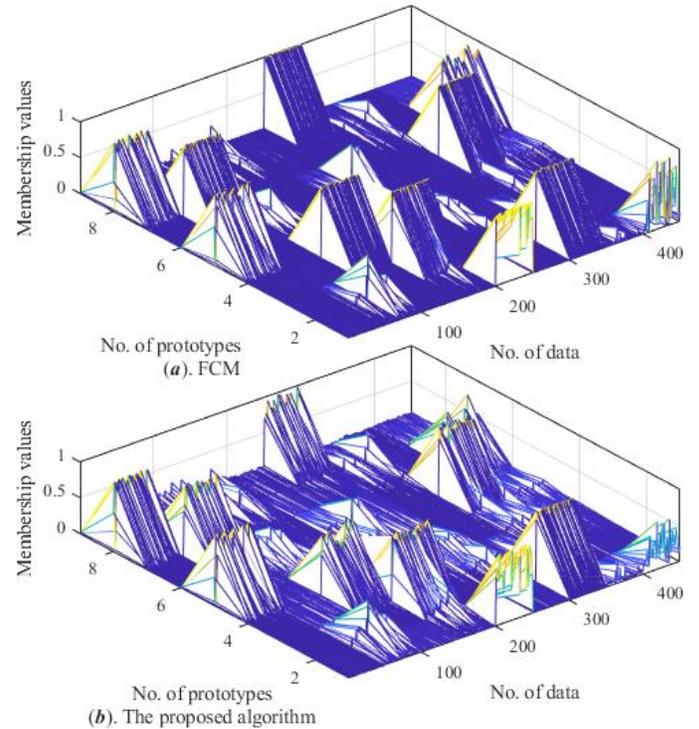

Fig. 6. Membership functions.

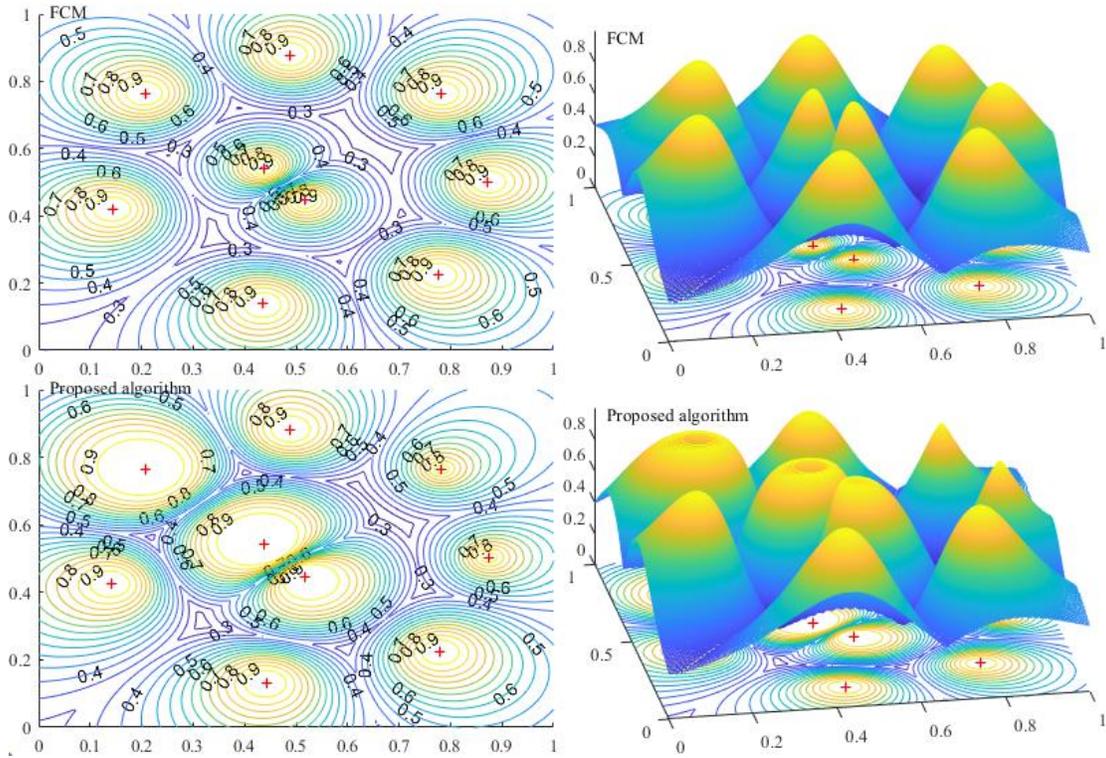

Fig. 7. Contour plots of membership functions.

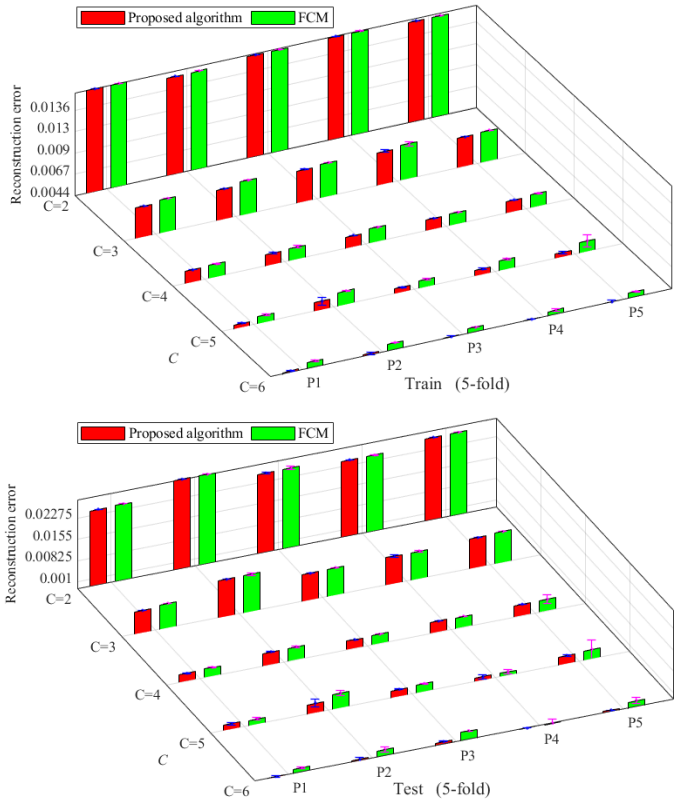

Fig. 8. Reconstruction errors of 5-fold cross validation with all the protocols for the synthetic dataset.

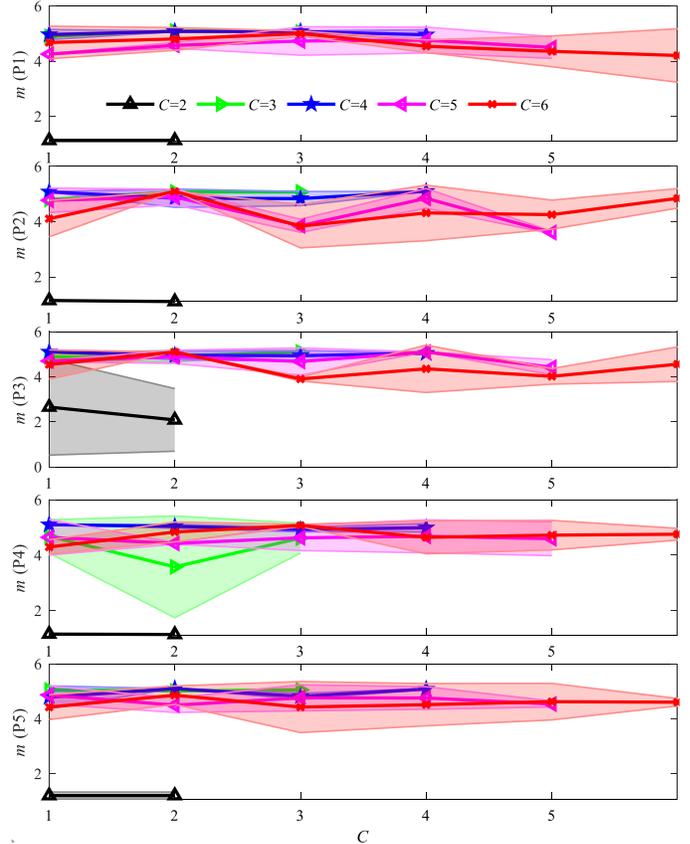

Fig. 9. Fuzzification factor vector of the proposed algorithm for the synthetic dataset.

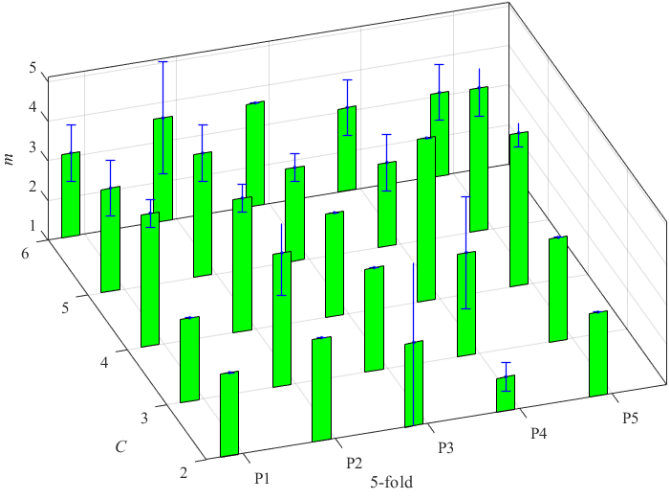

Fig. 10. Fuzzification factor of the FCM for the synthetic dataset.

*B. Publicly available data*

In this part, we report the results of reconstruction performance for eight publicly available datasets (see UCI machine learning repository for more details). The results of the reconstruction errors of 5-fold cross validation with all the protocols for the iris dataset and their corresponding fuzzification factors are plotted in Figs. 12–13. For the other publicly datasets we provide a list of the mean of the protocols which are summarized in Tables I–XII. It is noticeable that the reconstruction performance of all the datasets is improved with the use of the proposed approach, although the improvement of some datasets is not very significant. For several datasets the proposed scheme has significant advantages over the FCM algorithm, such as the iris, glass identification, statlog (heart) and connectionist bench datasets. For the wine and breast cancer datasets, the improvement is little, partially due to the dataset structure. In addition, as we have posted before, the FCM is a special case of the proposed approach.

In other words, the fuzzification factor helps control the shape of the clusters (membership functions) and produce a balance between the membership grades close to 0 or 1 and those with intermediate values. Assigning different values to the fuzzification factor for each cluster increases the flexibility of the method (location of the prototypes); thus, the degranulation can be improved by optimizing the prototypes (assigning a reasonable fuzzy factor to each prototype). Furthermore, the degranulation (reconstruction error) is a function of the fuzzification factor as well as the number of clusters. As such it can serve as a suitable measure to choose the optimal values of these parameters.

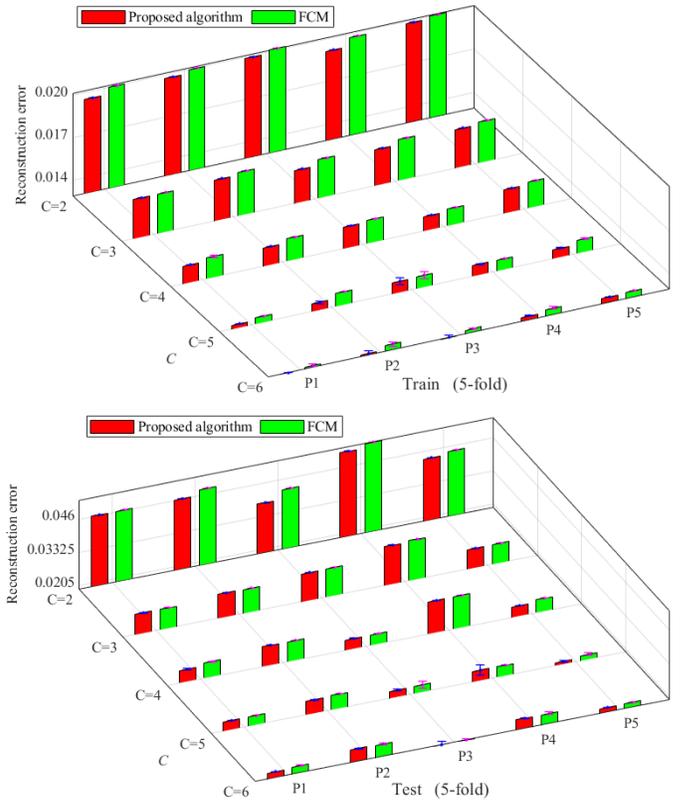

Fig. 11. Reconstruction errors of 5-fold cross validation with all the protocols for the iris dataset.

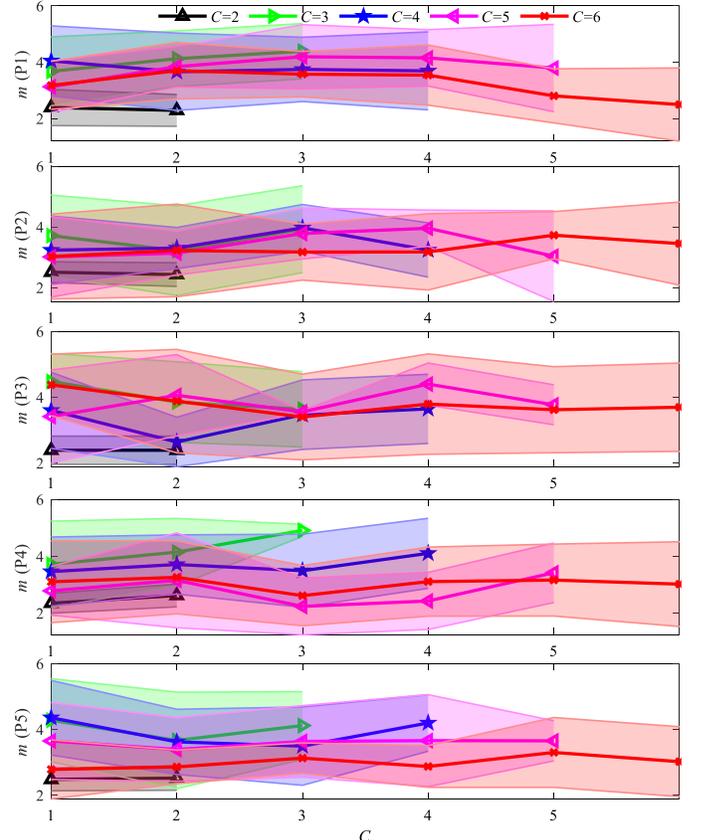

Fig. 12. Values of the fuzzification factor vector of the proposed algorithm for the iris dataset.

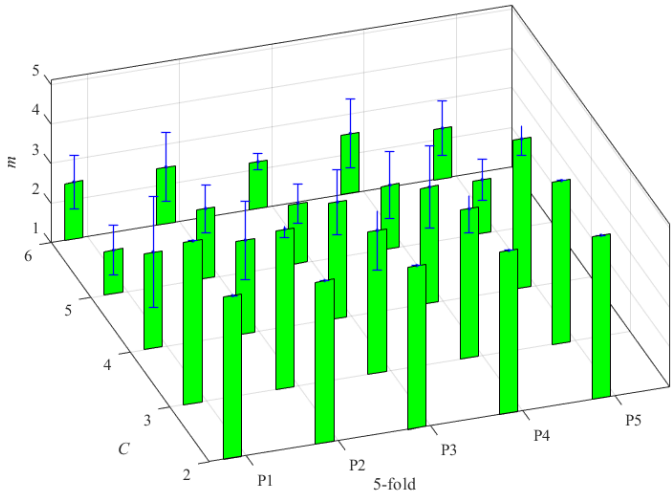

Fig. 13. Fuzzification factor of the FCM for the iris dataset.

Table I. Results of reconstruction error of the user dataset.

| Dataset | | User Knowledge Modeling | |
|---|---|---|---|
| C | Methods | FCM | Proposed method |
| 2 | Train | 0.0188 ± 0.0004 | 0.0186 ± 0.0004 |
| 3 | | 0.0170 ± 0.0004 | 0.0165 ± 0.0004 |
| 4 | | 0.0153 ± 0.0006 | 0.0152 ± 0.0005 |
| 5 | | 0.0143 ± 0.0002 | 0.0138 ± 0.0004 |
| 6 | | 0.0138 ± 0.0004 | 0.0131 ± 0.0003 |
| 2 | Test | 0.0410 ± 0.0017 | 0.0410 ± 0.0017 |
| 3 | | 0.0374 ± 0.0028 | 0.0371 ± 0.0033 |
| 4 | | 0.0352 ± 0.0030 | 0.0351 ± 0.0025 |
| 5 | | 0.0349 ± 0.0022 | 0.0348 ± 0.0017 |
| 6 | | 0.0335 ± 0.0032 | 0.0333 ± 0.0033 |
| 2 | Total | 0.0232 | 0.0231 |
| 3 | | 0.0211 | 0.0206 |
| 4 | | 0.0193 | 0.0192 |
| 5 | | 0.0184 | 0.0180 |
| 6 | | 0.0178 | 0.0172 |
| | Mean | 0.0200 | 0.0196 |
| 2 | m & $m$ | 1.20 | [1.77, 1.76] |
| 3 | | 2.10 | [2.22, 1.95, 2.12] |
| 4 | | 1.85 | [1.177, 1.16, 1.17, 1.17] |
| 5 | | 1.80 | [2.44, 2.60, 2.89, 2.49, 2.84] |
| 6 | | 1.80 | [2.98, 2.51, 2.05, 2.605, 2.63, 2.38] |

Table II. Results of reconstruction error of the glass identification dataset.

| Dataset | | Glass Identification | |
|---|---|---|---|
| C | Methods | FCM | Proposed method |
| 2 | Train | 0.0172 ± 0.0008 | 0.0168 ± 0.0009 |
| 3 | | 0.0175 ± 0.0009 | 0.0157 ± 0.0009 |
| 4 | | 0.0155 ± 0.0009 | 0.0144 ± 0.0007 |
| 5 | | 0.0129 ± 0.0004 | 0.0123 ± 0.0002 |
| 6 | | 0.0123 ± 0.0005 | 0.0112 ± 0.0005 |
| 2 | Test | 0.0417 ± 0.0083 | 0.0412 ± 0.0069 |
| 3 | | 0.0418 ± 0.0051 | 0.0375 ± 0.0052 |
| 4 | | 0.0391 ± 0.0052 | 0.0366 ± 0.0016 |
| 5 | | 0.0342 ± 0.0038 | 0.0341 ± 0.0046 |
| 6 | | 0.0323 ± 0.0030 | 0.0346 ± 0.0026 |
| 2 | Total | 0.0221 | 0.0217 |
| 3 | | 0.0224 | 0.0200 |
| 4 | | 0.0202 | 0.0189 |
| 5 | | 0.0171 | 0.0166 |
| 6 | | 0.0163 | 0.0159 |
| | Mean | 0.0196 | 0.0186 |
| 2 | m & $m$ | 1.95 | [1.13, 1.12] |
| 3 | | 1.45 | [3.22, 3.31, 2.92] |
| 4 | | 1.45 | [2.45, 1.86, 2.20, 2.11] |
| 5 | | 2.45 | [2.46, 1.89, 2.53, 2.42, 2.19] |
| 6 | | 1.65 | [1.48, 1.49, 1.43, 1.45, 1.50, 1.49] |

Table III. Results of reconstruction error of the statlog (heart) dataset.

| Dataset | | Statlog (Heart) | |
|---|---|---|---|
| C | Methods | FCM | Proposed method |
| 2 | Train | 0.0347 ± 0.0004 | 0.0345 ± 0.0004 |
| 3 | | 0.0326 ± 0.0005 | 0.0322 ± 0.0005 |
| 4 | | 0.0324 ± 0.0004 | 0.0320 ± 0.0004 |
| 5 | | 0.0321 ± 0.0004 | 0.0314 ± 0.0003 |
| 6 | | 0.0305 ± 0.0007 | 0.0289 ± 0.0010 |
| 2 | Test | 0.0740 ± 0.0040 | 0.0741 ± 0.0025 |
| 3 | | 0.0706 ± 0.0028 | 0.0715 ± 0.0027 |
| 4 | | 0.0709 ± 0.0027 | 0.0709 ± 0.0027 |
| 5 | | 0.0706 ± 0.0032 | 0.0703 ± 0.0061 |
| 6 | | 0.0707 ± 0.0036 | 0.0702 ± 0.0055 |
| 2 | Total | 0.0426 | 0.0425 |
| 3 | | 0.0402 | 0.0401 |
| 4 | | 0.0401 | 0.0397 |
| 5 | | 0.0398 | 0.0392 |
| 6 | | 0.0385 | 0.0372 |
| | Mean | 0.0402 | 0.0397 |
| 2 | m & $m$ | 1.60 | [3.39, 3.45] |
| 3 | | 1.90 | [4.86, 4.82, 4.75] |
| 4 | | 2.15 | [3.64, 3.98, 4.14, 2.97] |
| 5 | | 1.90 | [3.49, 3.58, 3.05, 3.55, 3.07] |
| 6 | | 1.80 | [3.10, 3.03, 3.27, 2.12, 3.04, 2.90] |

Table IV. Results of reconstruction error of the connectionist bench dataset.

| Dataset | | Connectionist Bench | |
|---|---|---|---|
| C | Methods | FCM | Proposed method |
| 2 | Train | 0.0569 ± 0.0011 | 0.0564 ± 0.0012 |
| 3 | | 0.0403 ± 0.0008 | 0.0389 ± 0.0007 |
| 4 | | 0.0391 ± 0.0006 | 0.0375 ± 0.0003 |
| 5 | | 0.0363 ± 0.0024 | 0.0345 ± 0.0024 |
| 6 | | 0.0335 ± 0.0015 | 0.0314 ± 0.0013 |
| 2 | Test | 0.1227 ± 0.0063 | 0.1225 ± 0.0071 |
| 3 | | 0.0969 ± 0.0072 | 0.0975 ± 0.0054 |
| 4 | | 0.0964 ± 0.0041 | 0.0947 ± 0.0095 |
| 5 | | 0.0928 ± 0.0074 | 0.0914 ± 0.0085 |
| 6 | | 0.0908 ± 0.0081 | 0.0914 ± 0.0083 |
| 2 | Total | 0.0701 | 0.0696 |
| 3 | | 0.0516 | 0.0506 |
| 4 | | 0.0506 | 0.0490 |
| 5 | | 0.0476 | 0.0459 |
| 6 | | 0.0450 | 0.0434 |
| Mean | | 0.0530 | 0.0517 |
| 2 | | 1.20 | [1.57, 1.46] |
| 3 | | 2.10 | [4.15, 4.26, 4.14] |
| 4 | m & *m* | 1.50 | [2.96, 2.87, 2.95, 2.82] |
| 5 | | 1.70 | [2.52, 2.47, 2.55, 2.54, 2.66] |
| 6 | | 1.75 | [3.42, 3.267, 3.37, 3.14, 3.15, 3.43] |

Table IV. Results of reconstruction error of the wine dataset.

| Dataset | | Wine | |
|---|---|---|---|
| $C$ | Methods | FCM | Proposed method |
| 2 | | 0.0233 ± 0.0003 | 0.0230 ± 0.0003 |
| 3 | | 0.0182 ± 0.0004 | 0.0181 ± 0.0004 |
| 4 | Train | 0.0175 ± 0.0004 | 0.0171 ± 0.0004 |
| 5 | | 0.0167 ± 0.0005 | 0.0162 ± 0.0004 |
| 6 | | 0.0159 ± 0.0004 | 0.0154 ± 0.0004 |
| 2 | | 0.0515 ± 0.0048 | 0.0510 ± 0.0053 |
| 3 | | 0.0430 ± 0.0035 | 0.0432 ± 0.0062 |
| 4 | Test | 0.0429 ± 0.0034 | 0.0425 ± 0.0036 |
| 5 | | 0.0415 ± 0.0037 | 0.0416 ± 0.0031 |
| 6 | | 0.0406 ± 0.0034 | 0.0403 ± 0.0038 |
| 2 | | 0.0290 | 0.0286 |
| 3 | | 0.0232 | 0.0232 |
| 4 | Total | 0.0226 | 0.0222 |
| 5 | | 0.0216 | 0.0213 |
| 6 | | 0.0209 | 0.0204 |
| Mean | | 0.0234 | 0.0231 |
| 2 | | 2.00 | [1.12, 1.13] |
| 3 | | 2.40 | [1.12, 1.16, 1.12] |
| 4 | m & *m* | 1.70 | [1.73, 1.58, 1.56, 1.52] |
| 5 | | 1.25 | [1.75, 1.76, 1.73, 1.82, 1.77] |
| 6 | | 1.45 | [1.52, 1.43, 1.42, 1.66, 1.56, 1.58] |

Table V. Results of reconstruction error of the breast cancer dataset.

| Dataset | | Breast Cancer | |
|---|---|---|---|
| $C$ | Methods | FCM | Proposed method |
| 2 | | 0.0337 ± 0.0001 | 0.0337 ± 0.0001 |
| 3 | | 0.0329 ± 0.0006 | 0.0328 ± 0.0006 |
| 4 | Train | 0.0265 ± 0.0005 | 0.0244 ± 0.0006 |
| 5 | | 0.0246 ± 0.0004 | 0.0234 ± 0.0001 |
| 6 | | 0.0236 ± 0.0005 | 0.0220 ± 0.0005 |
| 2 | | 0.0704 ± 0.0002 | 0.0701 ± 0.0031 |
| 3 | | 0.0690 ± 0.0012 | 0.0707 ± 0.0010 |
| 4 | Test | 0.0592 ± 0.0030 | 0.0578 ± 0.0031 |
| 5 | | 0.0571 ± 0.0034 | 0.0562 ± 0.0022 |
| 6 | | 0.0528 ± 0.0042 | 0.0540 ± 0.0026 |
| 2 | | 0.0410 | 0.0409 |
| 3 | | 0.0401 | 0.0404 |
| 4 | Total | 0.0330 | 0.0311 |
| 5 | | 0.0311 | 0.0300 |
| 6 | | 0.0294 | 0.0284 |
| Mean | | 0.0349 | 0.0342 |
| 2 | m & *m* | 2.15 | [2.83, 2.99] |
| 3 | | 5.00 | [4.59, 4.17, 4.49] |
| 4 | | 2.30 | [3.61, 3.97, 3.35, 3.35] |
| 5 | | 3.50 | [2.50, 2.01, 2.55, 2.18, 2.66] |
| 6 | | 3.20 | [1.72, 1.58, 1.82, 1.94, 1.74, 1.80] |

Table VI. Results of reconstruction error of the buddy move dataset.

| Dataset | | Buddy Move | |
|---|---|---|---|
| $C$ | Methods | FCM | Proposed method |
| 2 | | 0.0187 ± 0.0002 | 0.0186 ± 0.0002 |
| 3 | | 0.0130 ± 0.0003 | 0.0129 ± 0.0003 |
| 4 | Train | 0.0124 ± 0.0005 | 0.0120 ± 0.0005 |
| 5 | | 0.0110 ± 0.0004 | 0.0105 ± 0.0003 |
| 6 | | 0.0100 ± 0.0004 | 0.0092 ± 0.0003 |
| 2 | | 0.0383 ± 0.0017 | 0.0383 ± 0.0018 |
| 3 | | 0.0283 ± 0.0020 | 0.0292 ± 0.0021 |
| 4 | Test | 0.0274 ± 0.0021 | 0.0273 ± 0.0026 |
| 5 | | 0.0258 ± 0.0027 | 0.0253 ± 0.0028 |
| 6 | | 0.0222 ± 0.0009 | 0.0221 ± 0.0010 |
| 2 | | 0.0226 | 0.0226 |
| 3 | | 0.0160 | 0.0162 |
| 4 | Total | 0.0154 | 0.0151 |
| 5 | | 0.0139 | 0.0134 |
| 6 | | 0.0124 | 0.0118 |
| Mean | | 0.0161 | 0.0158 |
| 2 | | 1.90 | [1.83, 1.61] |
| 3 | | 4.60 | [3.81, 3.52, 4.21] |
| 4 | m & *m* | 3.90 | [3.42, 3.61, 3.75, 3.50] |
| 5 | | 3.30 | [4.059, 3.95, 3.83, 4.31, 3.92] |
| 6 | | 2.70 | [2.83, 3.90, 3.99, 3.92, 3.65, 2.82] |

## V. CONCLUSIONS

In this study, we develop an enhanced scheme of the degranulation mechanism. During the design process, we define a vector of fuzzification factor to assign an appropriate fuzzification factor for each prototype. With the supervised learning mode of the granulation-degranulation, the PSO is used to optimize the entries of the vector of fuzzification factor to obtain optimal the prototypes and the partition matrix that ultimately enhance the performance of the degranulation mechanism. We carry out a comprehensive analysis and provide a series of experiments. Both of them demonstrate the effectiveness of the proposed scheme. To the best of our knowledge, this research scheme is exposed for the first time. We show that the algorithm can enhance the performance of the degranulation mechanism. Unfortunately, the proposed method involves the PSO optimization, leading to some additional computing overhead.

The proposed models open a specific way for enhancing the performance of the degranulation mechanism and pose a more general problem concerning the reduction of computational complexity. Future work also includes the study of the deep relationship between the degranulation error and the classification rate.